# VCVW-3D: A Virtual Construction Vehicles and Workers Dataset with 3D Annotations

Yuexiong Ding[a,b], Xiaowei Luo[a,b*]

[a]Department of Architecture and Civil Engineering, City University of Hong Kong, Hong Kong, China

[b]Architecture and Civil Engineering Research Center, Shenzhen Research Institute of City University of Hong Kong, Shenzhen, China

**Abstract**

Currently, object detection applications in construction are almost based on pure 2D data (both image and annotation are 2D-based), resulting in the developed artificial intelligence (AI) applications only applicable to some scenarios that only require 2D information. However, most advanced applications usually require AI agents to perceive 3D spatial information, which limits the further development of the current computer vision (CV) in construction. The lack of 3D annotated datasets for construction object detection worsens the situation. Therefore, this study creates and releases a virtual dataset with 3D annotations named VCVW-3D, which covers 15 construction scenes and involves ten categories of construction vehicles and workers. The VCVW-3D dataset is characterized by multi-scene, multi-category, multi-randomness, multi-viewpoint, multi-annotation, and binocular vision. Several typical 2D and monocular 3D object detection models are then trained and evaluated on the VCVW-3D dataset to provide a benchmark for subsequent research. The VCVW-3D is expected to bring considerable economic benefits and practical significance by reducing the costs of data construction, prototype development, and exploration of space-awareness applications, thus promoting the development of CV in construction, especially those of 3D applications.

# 1. Introduction

As one of the labor-intensive industries, construction still relies mainly on the human force. Thanks to the rapid development of artificial intelligence (AI) and computer vision (CV) technologies, automatic construction management based on cameras has become possible and attracted extensive attention from academia and industry. Various studies have been conducted on many aspects of construction [1], such as safety management, progress monitoring, quality control, and productivity tracking, significantly improving management and production efficiency.

As one of the critical tasks of CV, object detection plays an essential role on the road of construction automation [2–4]. However, current object detection research and applications in construction mainly focus on pure 2D data (both image and annotation are 2D-based), such as 2D object detection and 2D semantic/instance segmentation. Based only on 2D annotation information, the developed detection models lack spatial awareness. These models are only used in simple scenarios that do not require space information, seriously limiting CV's further promotion and application on construction sites. In fact, many critical management activities need to perceive the surrounding space of the target object, such as the judgment of high-altitude or edge operations, the space distance between large construction vehicles and workers, the relative position between the safety belt hook and the worker, the distance between the crane cargo and surrounding stuff, etc. Therefore, the developed detection models must have the 3D perception ability to achieve advanced automation in construction management.

On the other hand, training a deep learning-based CV model requires a large amount of labeled data [5]. The complexity and dynamics of construction activities bring challenges to data acquisition and annotation. For example, Duan et al. [6] spent more than 1,800 labor hours processing, annotating, and reviewing only 2D bounding boxes for 21,863 construction images. The acquisition and annotation of 3D data are times more complicated and costly than 2D data, which is one of the main reasons why there is still no sizeable public dataset with 3D annotations in construction for object detection.

Therefore, this study constructs a **V**irtual **C**onstruction **V**ehicles and **W**orkers dataset with 3D annotations (VCVW-3D) based on the Unity 3D platform. By generating and annotating data in a controllable and customizable way on the programmable virtual platform, the difficulties of 3D data acquisition and labeling mentioned above can be significantly alleviated [7,8]. The VCVW-3D dataset is characterized by multi-scene (15 scenes indoor and outdoor), multi-category (224 prefabricated assets in 10 categories), multi-randomnesses (random number, space position, angle, and color of the object and random global lighting), multi-viewpoint, multi-annotation (2D/3D bounding box, 2D semantic/instance segmentation, depth map), and binocular stereo vision. A benchmark of 2D and monocular 3D object detection on the VCVW-3D is also provided for reference. The VCVW-3D might bring considerable economic benefits by reducing data construction and prototype development costs. Besides, more and more 3D CV research will become explorable, thus promoting the development of CV in construction, especially those of 3D object detection applications.

The rest of the paper is organized as follows: section 2 introduces related works of object detection and relevant datasets. Section 3 elaborates on constructing the VCVW-3D dataset, while section 4 conducts data preview and statistics. Section 5 provides a benchmark of 2D/3D object detection on the VCVW-3D. The contributions, limitations, and future work of this work are further discussed in section 6. Finally, section 7 concludes the study.

## 2. Literature review

### 2.1 Deep neural network-based object detection

2D object detection can be divided into two-stage and single-stage object detection. The two-stage models, also known as the region-based methods, extract a series of bounding boxes in the first stage and then use convolutional neural networks (CNN) for classification. Two-stage detection models have higher accuracy but relatively slow inference speeds. The Region-CNN (R-CNN) [9], Faster R-CNN [10], and Mask R-CNN [11] are some classical two-stage detection models. One-stage methods directly predict results from the extracted features to improve the inference speed. Typical one-stage models are You Only Look Once (YOLOs) [12–15] and Single Shot Detector (SSD) [16].

3D object detection based on monocular or binocular 2D images is an emerging research hotspot, which can estimate objects' 3D space information economically by only using the information from the 2D image. Typical methods include monocular and binocular stereo 3D object detection. Currently, some advanced monocular 3D object detection models include SMOKE [17], FCOS3D [18], PGD [19], MonoFlex [20], ImVoxelNet [21], while binocular/stereo models are Disp R-CNN [22], DSGN [23], YoLoStereo3D [24], DSGN++ [25], etc.

**2.2 Object detection in construction**

Traditional 2D object detection generally refers to bounding box-based detection, while semantic or instance segmentation can be regarded as pixel-level object detection. Both traditional and extended object detection has been widely studied and applied in many aspects of construction [2–4,26–28]. For example, Fang et al. [4] developed an intelligent model for detecting workers' safety belts based on Faster R-CNN networks. Chen et al. [2] analyzed the production efficiency of the excavator according to multiple consecutive frames using object detection and tracking technologies. Panella et al. [27] trained a U-Net model for crack detection and segmentation, while Kang et al. [28] developed a YOLACT model for site object instance segmentation. However, these models can only be used to determine whether the object exists but cannot access the object's spatial information, which limits their further application and promotion on construction sites.

Currently, most 3D object detection applications in construction focus on point cloud processing, i.e., semantic/instance segmentation of 3D point clouds [29–32]. For example, Yin et al. [30] trained a PointNet model for the semantic segmentation of indoor point clouds. Similarly, Chew et al. [32] developed an encoder-decoder model to segment large-scale urban and rural 3D point clouds. However, 3D point cloud data only contains position information but loses the visual information of the scene. Only a few studies focus on bounding box-related 3D applications in construction [33,34]. For example, Wang et al. [33] developed a robot with spatial perception for construction waste recycling by combining 2D object detection and depth awareness of LiDAR. Yan's research [34] was the first research to conduct 3D boundary box prediction on 2D images, which estimated 3D boundary boxes according to the detected 2D bounding boxes of the whole, side, and front/back of construction vehicles. However, these two studies were not complete 3D object detection since they both applied 2D object detection models actually.

**2.3 Datasets for object detection**

There are many large public 2D datasets for object detection in the general area, such as PASCAL VOC [35], COCO [36], and ILSVRC [37]. The current development of 3D object detection is mainly driven by autonomous driving research. Many large 3D datasets in this field thus have been publicly released, such as KITTI [38], nuScenes [39], Waymo [40], Lyft Level5 [41], etc. These public datasets have spawned a series of excellent 3D CV models, greatly promoting the development of 3D CV technologies.

Unlike the data isolation problem existing in other aspects of construction [41], an increasing number of datasets for object detection research have been chosen to be released publicly. For example, Roberts et al. [42] published a dataset containing 479 atomic videos of excavators and dump trucks for detecting, tracking, and activity analysis of earthmoving equipment. Xuehui et al. [43] created the MOCS dataset by collecting 41,668 images from 174 construction sites with annotations of 13 types of objects. Xiao et al. [44] manually collected and annotated 10,000 images of 10 types of construction machines to construct the ACID dataset for construction machinery recognition. Duan et al. [6] spent more than 1,800 labor hours building and labeling the SODA dataset of 21,863 construction images of 15 categories in worker, material, machine, and layout. However, these datasets were only labeled with 2D information and thus useless to 3D reserarch.

As mentioned above, most of the 3D object detection research in construction focused on point cloud segmentation. Unfortunately, only a few of those researchers choose to publish their data. For example, Hackel et al. [45] published a 3D point cloud dataset with eight categories covering 30 urban outdoor scenes. However, there are no 3D bounding box annotations for related objects in point cloud datasets. From this perspective, few accessible 3D annotated data for reference is one of the main reasons for the lack of 3D object detection research in the current construction field. Referring to the rapid development experience of 3D object detection in autonomous driving, it is necessary to construct and release a dataset for 3D object detection to promote the further development of 3D CV in construction.

## 3. Methodology

As shown in Figure 1, the construction of the VCVW-3D dataset mainly includes the following steps: 1) objects and scenes creation and collection, 2) image synthesis and 3) data annotation. The data generation and annotation programs were developed based on the Unity 3D platform and its perception packages [46,47].

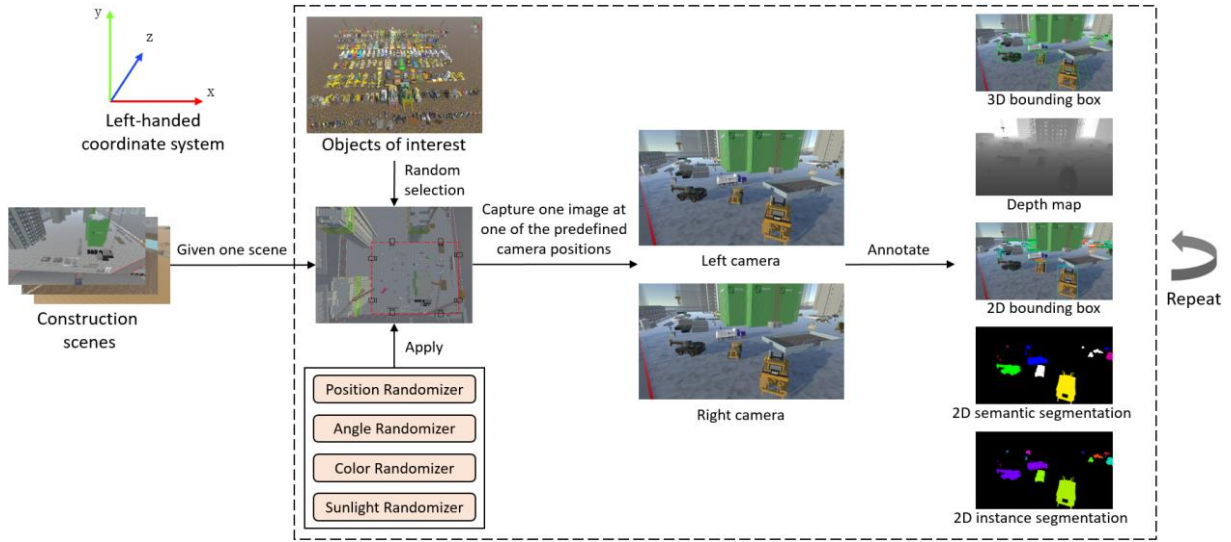

Figure 1. Process of data synthesis and annotation.

**3.1 Object categories**

*Basic information*. The VCVW-3D dataset involved ten object categories: workers, dump trucks, excavators, concrete mixers, forklifts, cranes (excluding tower cranes), loaders, bulldozers, graders, and road rollers. Multiple prefabricated assets were built for each category in Unity to achieve diversity in the same category, as shown in Figure 2. The number of prefabricated assets in each category is counted in Table 1. There were a total of 224 prefabricated assets in the category of interest, of which the worker had the most assets (41) while the grader was the least one (10). In addition, an extra 50 kinds of construction-related stuff were also involved in simulating the site conditions as naturally as possible, including wooden boxes, cement pipes, muck/brick piles, oil drums, roadblocks, etc.

*Orientation definition of prefabricated assets*. The orientation of the object is of great significance in 3D position annotation. This study defines the orientation of the asset model as the negative direction of the z-axis of its local coordinates. For the asset model without rotation joints, the orientation is consistent with its head orientation, as shown in Figure 3 (d). However, the orientation might have diverse definitions when the model has rotation joints. For example, the neck and waist of the worker and the chassis of the excavator and crane are rotatable. In daily construction activities, more attention is generally attached to the worker's body trunk since it is the largest part of the human body, while more attention focuses on the arms of the excavator and crane because their other components are almost immobile when they are working. Therefore, the orientation of the worker was determined as the orientation of its body trunk, and the orientations of the excavator and crane followed the orientation of their arms, as shown in (a), (b), and (c) of Figure 3. Models of other construction stuff did not define orientations since they were not the category of interest.

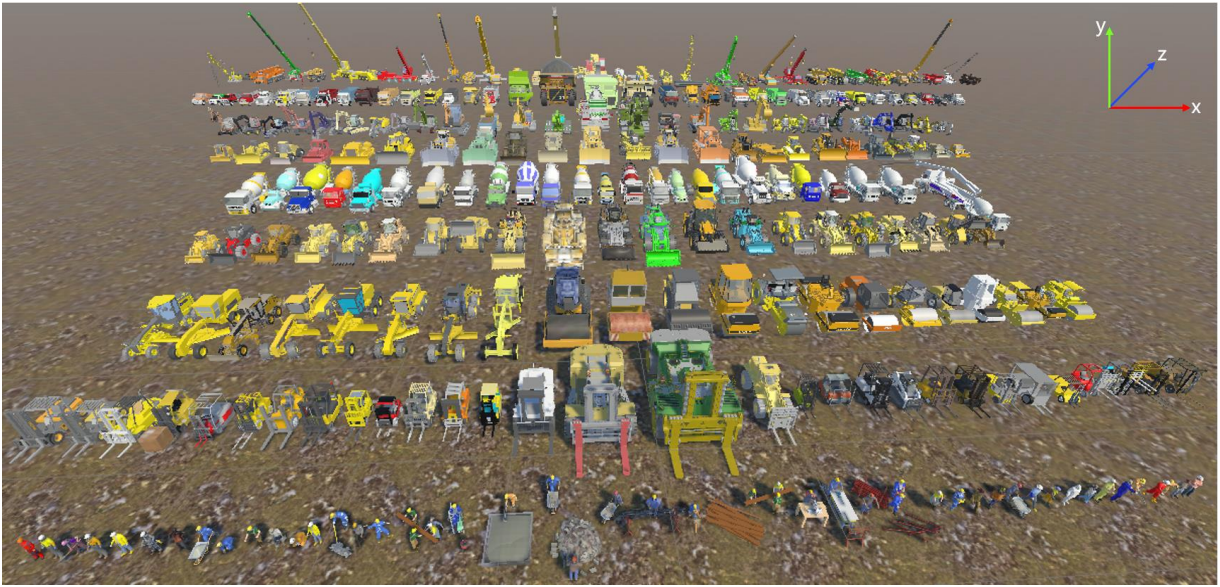

Figure 2. Prefabricated assets of the category of interest.

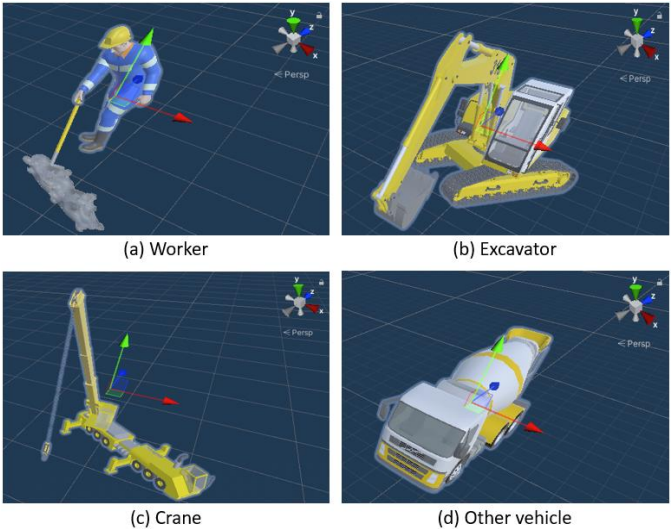

Figure 3. Models' orientation definition based on the left-handed coordinate system.

*Data split of prefabricated assets*. The prefabricated assets of the category of interest were divided into the training&validation (Trainval) and test sets with the ratio of [0.8: 0.2] before data synthesis to ensure the object instances in the Trainval set do not repeatedly appear in the test phase. Since construction stuff assets were not the category of interest, the Trainval and test set involved all those assets.

Table 1. The number of prefabricated assets in different sets.

| Categories | Number | | |
|---|---|---|---|
| | Total | Trainval | Test |
| Worker | 41 | 33 | 8 |
| Dump truck | 36 | 29 | 7 |
| Excavator | 29 | 23 | 6 |
| Concrete mixer | 25 | 20 | 5 |
| Forklift | 30 | 24 | 6 |
| Crane | 35 | 28 | 7 |
| Loader | 21 | 17 | 4 |
| Bulldozer | 25 | 20 | 5 |
| Grader | 10 | 8 | 2 |
| Road roller | 13 | 10 | 3 |
| Other construction stuff | | 50 | |

## 3.2 Construction scenes

The surroundings in different construction activities somehow influence the model's detection accuracy. Therefore, 15 different virtual indoor and outdoor scenes were constructed to cover as many construction activities as possible, such as concrete pouring, rebar binding, road laying, earthwork excavation, roof construction, indoor construction, and other common operations. Figure 4 shows the snapshots of the selected scenes. To be more consistent with the facts, different construction scenes were restricted to involving only specific object categories, as shown in Table 2. For example, indoor scenes only included workers, small forklifts, and small construction stuff. In outdoor high-altitude construction scenarios, workers, forklifts, and construction stuff were included, while excavators could also appear in particular cases. The outdoor construction scenes on the ground included all categories in Table 1. In addition, some construction scenes had multiple working floors, and the possible object categories between different floors were also different (e.g., scenes M and N in Figure 4).

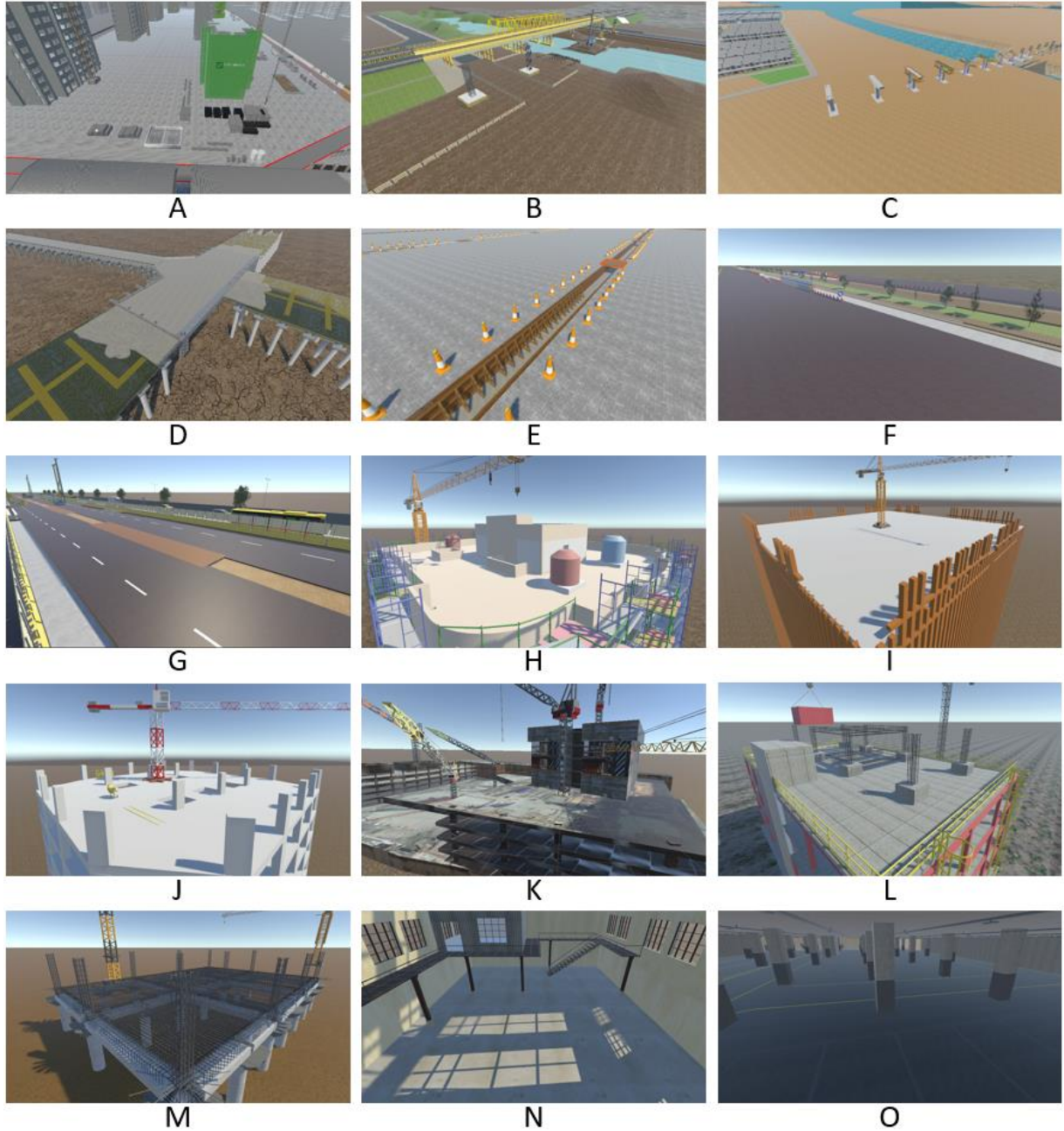

Figure 4. Snapshots of the selected construction scenes.

Table 2. Detailed restrictions and settings for each construction scene.

| Scenarios | Scenario types | Involved objects | Camera angle on the x-axis (degree)[*] | Camera high (meter)[*] |
|---|---|---|---|---|
| A | General | All | 30 | 10 |
| B | Bridge/road | All | 20 | 12 |
| C | Bridge/road | All | 20 | 10 |
| D | Bridge/road | All | 20 | 10 |
| E | Road | All | 20 | 9 |
| F | Road | All | 20 | 5.4 |
| G | Road | All | 20 | 5.6 |
| H | Roof | Workers, forklifts, small construction stuff | 10 | 5 |
| I | Roof | Workers, forklifts, excavators, construction stuff | 20 | 10 |
| J | Roof | Workers, forklifts, excavators, construction stuff | 10 | 5 |
| K | Roof | Workers, forklifts, excavators, construction stuff | 10 | 10 |
| L | Roof | Workers, forklifts, small construction stuff | 20 | 5 |
| M | Roof | Ground floor: All<br>First floor: Workers | 20 | 10 |
| N | Indoor | Ground floor: Workers, forklifts, construction stuff<br>First floor: Workers | 30 | 9 |
| O | Indoor | Workers, forklifts, construction stuff | 30 | 2.7 |

\* The optimal camera layout is out of the research scope of this study. Therefore, the camera high and angle were predetermined based on experience.

## 3.3 Image synthesis

When a scene was specified, a series of images were synthesized by repeating the following three steps:

- First, a random number of prefabricated assets were selected from the model library (restricted in Table 2) by using the sampling with replacement method. The range of the asset model number was customized before image synthesis based on the scene's monitoring area size. Each prefabricated asset had the same probability of being selected. The same asset model could be selected multiple times to generate different object instances in the same category.

- Then, multiple randomizers related to the position, angle, color, and sunlight, were applied to change the attributes of the generated object instances and the light source. The position, angle, and color randomizers acted on each object instance to achieve rich object diversity. The ranges of instances' random positions in different scenes were diverse and predefined before image synthesis, as red dashed rectangular boxes shown in Figure 5. The angle randomizer only rotated the instance around its y-axis. To avoid crossing/collision, a verification program was applied to each instance before putting it into the scene to check whether the current instance collides with other placed instances or static scene objects (e.g., the walls). In case of any collision, the current instance would be destroyed immediately. The sunlight randomizer simulated the sunlight randomly from the range of different latitudes (N90° ~ S90°), days of the year (1~365), and day times (6:00 am ~ 6:00 pm).

- Finally, put a pair of cameras into a specific position to render and capture the current scene. The two cameras were 0.3 meters apart and worked independently of each other. For different sizes of the monitored areas, the candidate camera positions were also different. As shown in Figure 5, there were mainly three camera deployment schemes considered in this study (not optimal but only for multi-viewpoints capture), which were applied to small areas, large square-like areas, and large rectangle areas, respectively. The camera height and rotation angle in different scenes were also predetermined, referring to Table 2. Note that image capture and data annotation were in processing simultaneously. The captured 2D visual information was saved as the 2D image, while the other information was used for data annotation, described in detail in section 3.4.

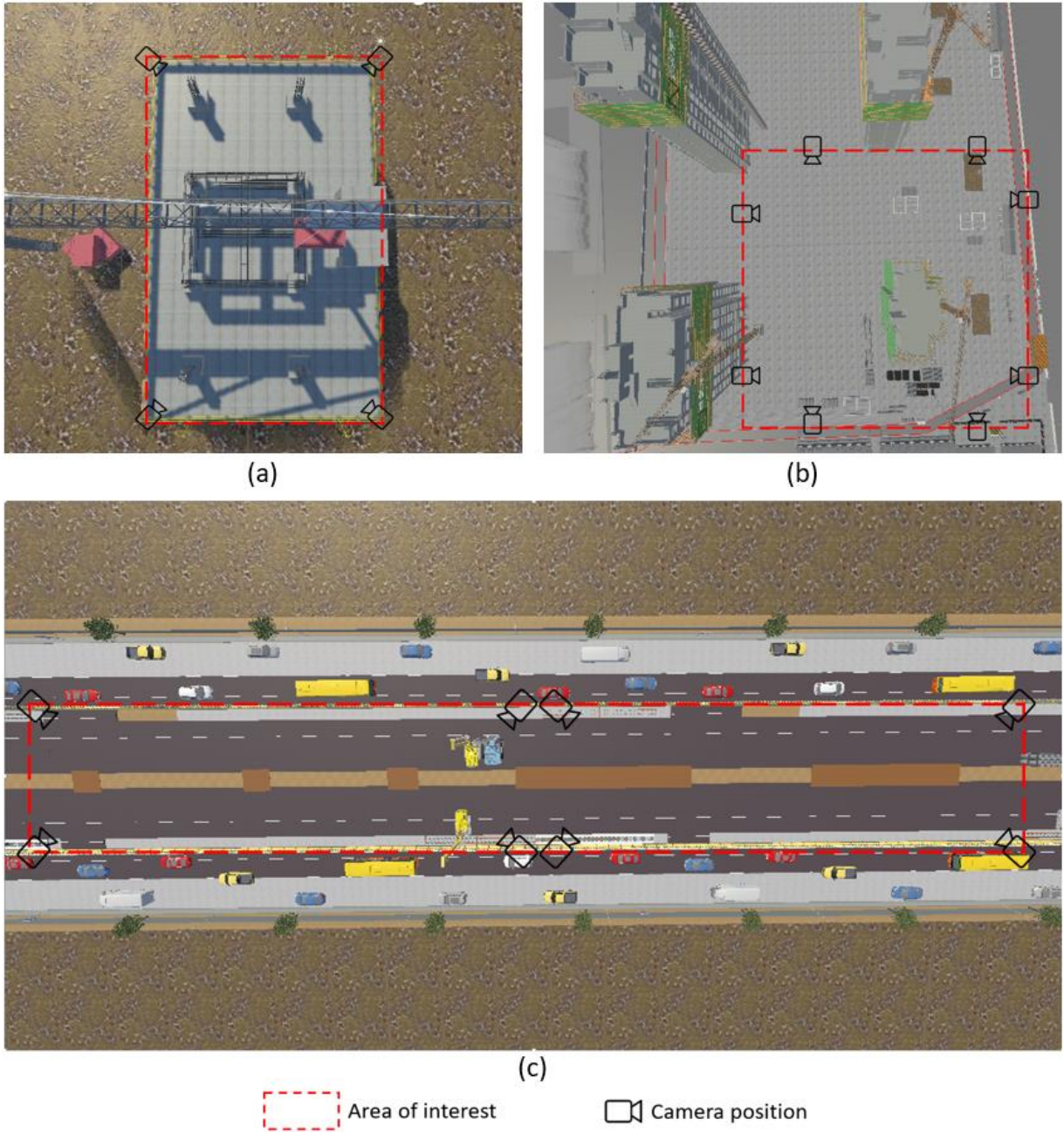

Figure 5. Candidate camera positions for different areas of interest.

**3.4 Data annotation**

The data annotation was simultaneous with image capturing. As shown in Table 3, each captured image had an annotation record, including the record ID, sensor information, image file path, and specific annotation information for different CV tasks. The main fields in the sensor item included the camera's translation, rotation, and intrinsic parameters, as shown in Table 4. The labeled information was different for different CV tasks, as shown in Table 5. For example, the 3D object detection task annotated the label ID, label name, translation, 3D size, and rotation for each object instance. The sum of visible pixels of each instance in the 2D image was also recorded. Note that the position and rotation of the camera were based on the global coordinate system, while the annotated 3D bounding boxes were based on the camera coordinate system. In this study, the global coordinate system was set as the translation of (0, 0, 0) and rotation of (1, 0, 0, 0).

Table 3. Main fields of one annotation record for each image.

```
"captures": [
  {
    "id":           <str>  --  Unique record ID.
    "seq_id":       <str>  --  Sequence ID for binocular vision annotation, indicating which two item
                               records were captured at the same timestamp.
    "sensor":       <obj>  --  Details of the sensor, refer to Table 4.
    "filename":     <str>  --  Filename of the captured image (e.g., 001.png).
    "annotations":  [<obj>, ...]  --  Annotations of this capture, refer to Table 5.
  },
  {…}
]
```

Table 4. Main fields of the "sensor" item.

```
"sensor ": {
  "sensor_id":        <str>  --  Sensor ID.
  "translation":      <float, float, float>  --  Sensor position(x, y, z) in meters, with respect to the global
                                                 coordinate system.
  "rotation":         <float, float, float, float>  --  Quaternion (w, x, y, z) of sensor orientation, with respect
                                                       to the global coordinate system.
  "camera_intrinsic": <3x3 float matrix>  --  Intrinsic camera calibration.
}
```

Table 5. Main fields of annotations for different tasks.

| Tasks | Annotations |
| --- | --- |
| 3D bounding box | {<br>  "id":                    <str> -- ID, indicating current task (e.g., 3D bounding box).<br>  "valuse": [<br>    {<br>      "label_id":    <int> -- Integer ID of the label.<br>      "label_name": <str> -- String label name.<br>      "instance_id": <int> -- Unique ID for each object instance.<br>      "translation": <float, float, float> -- Center location of the 3D bounding box<br>                                      in meters, with respect to the sensor's coordinate system. |
| | "size": <float, float, float> -- Width, height, and length of the 3D bounding box in meters.<br>"rotation": <float, float, float, float> -- Orientation (w, x, y, z) of the 3D bounding box, with respect to the sensor's coordinate system.<br>"visible_pixels": <float> -- The sum of visible pixels of the instance in the captured 2D image.<br>},<br>{...}<br>]<br>} |
| Depth map | {<br>"id": <str> -- ID, indicating current task (e.g., Depth map).<br>"filename": <str> -- Filename of the 16-bit depth map (e.g., depth_map.raw).<br>} |
| 2D bounding box | {<br>"id": <str> -- ID, indicateing current task (e.g., 2D bounding box).<br>"valuse": [<br>{<br>"label_id": <int> -- Integer ID of the label.<br>"label_name": <str> -- String label name.<br>"instance_id": <int> -- Unique ID for each object instance.<br>"x": <float > -- Upper left corner's x coordinate of the 2D bounding box.<br>"y": <float> -- Upper left corner's y coordinate of the 2D bounding box.<br>"width": <float> -- Width of the 2D bounding box in pixels.<br>"height": <float> -- Height of the 2D bounding box in pixels.<br>"visible_pixels": <float> -- The sum of visible pixels of the instance in the captured 2D image.<br>},<br>{...}<br>]<br>} |
| 2D semantic segmentation | {<br>"id": <str> -- ID, indicating current task (e.g., Sementic segmentation).<br>"filename": <str> -- Filename of the color mask image (e.g., segmentation.png).<br>} |
| 2D instance segmentation | {<br>"id": <str> -- ID, indicating current task (e.g., Instance segmentation).<br>"filename": <str> -- Filename of the color mask image (e.g., instance.png).<br>"valuse": [<br>{<br>"label_id": <int> -- Integer ID of the label.<br>"label_name": <str> -- String label name.<br>"instance_id": <int> -- Unique ID for each object instance.<br>"color": <float, float, float, float> -- Value from 0 to 1 for the red, green, blue, alpha channels to mask the instance. |

```
                    "visible_pixels": <float>  -- The sum of visible pixels of the instance in the
                                                   captured 2D image.
                },
                {…}
            ]
        }
```

## 4. Data preview and statistics

The VCVW-3D consisted of the training&validataion (Trainval) and test sets, which were generated according to prefabricated assets split results (Table 1) and scenes' restrictions and settings (Table 2). Each scene contributed 20,000 and 5,000 images of 1920×1080 for the Trainval and test sets, respectively. The data amount and image size can also be customized in other specific settings. Figure 6 shows sample images with annotations. With the help of multiple randomizers and camera poses, various object instances with different numbers, sizes, positions, and orientations were generated under different construction scenes and light conditions. The diversity of the object of interest thus was greatly improved. Multiple construction scenes further expanded the data representation domain of the VCVW-3D. There were also kinds of occlusions in the VCVW-3D, such as the occlusion caused by the scene assets (e.g., columns, walls, reinforcement, tower crane, etc.) and the mutual occlusion between the generated instances, which covered many common occlusion situations in the real construction scene.

In addition to the 3D bounding box, binocular stereo vision and depth map were the other two 3D-related attributes of the VCVW-3D, as shown in Figure 7. The space information under the current camera viewpoint was stored in the depth map, which can be reconstructed to the ground-truth 3D point cloud of the current scene, as shown in Figure 7 (d). Many 3D CV research in construction thus can be further studied based on the annotated binocular vision and depth map data, such as binocular 3D object detection, multimodal (2D image + depth map) 3D object detection, monocular/binocular 3D reconstruction, monocular/binocular depth estimation, etc.

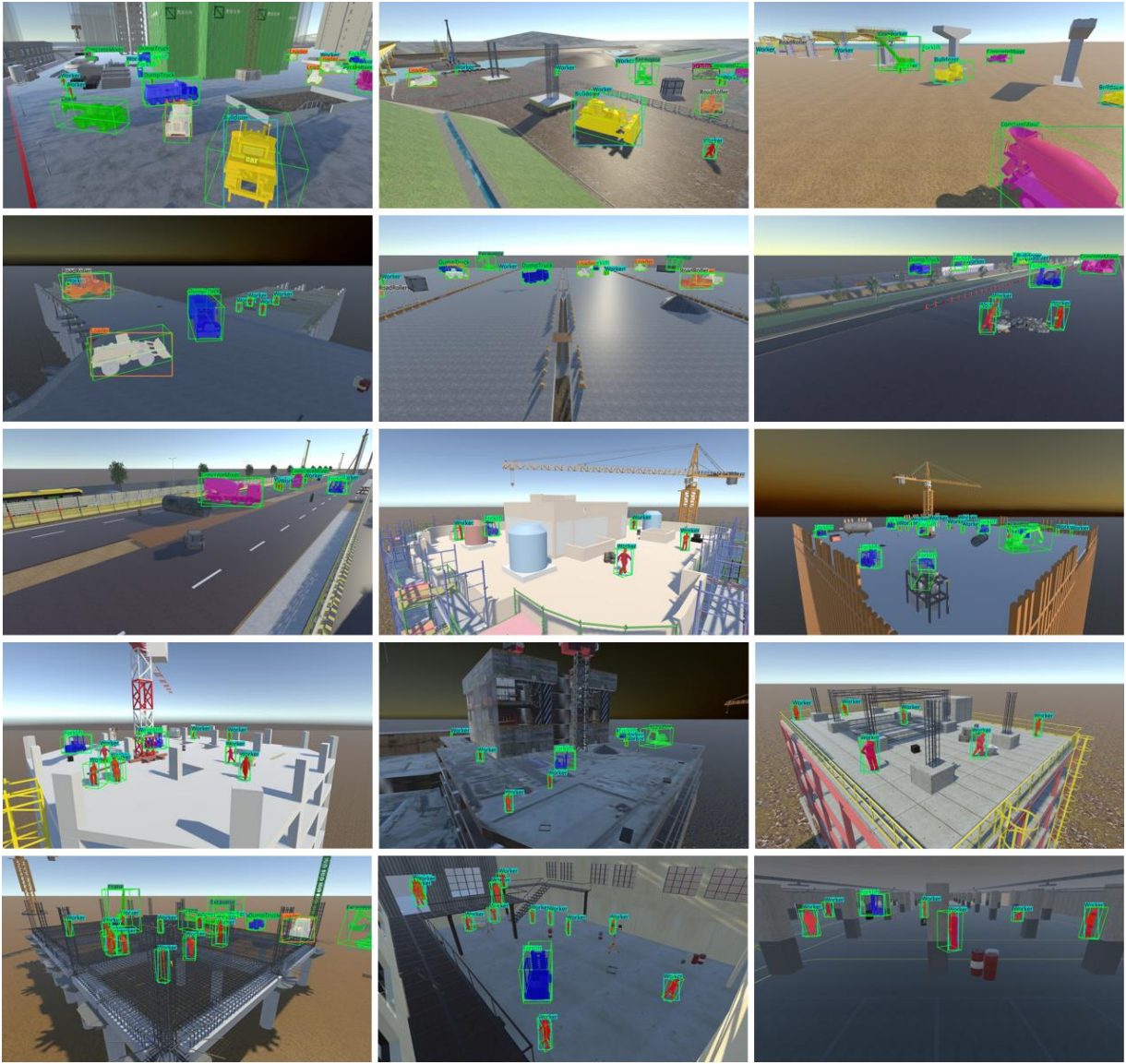

Figure 6. Sample images with annotations (3D/2D bounding box and semantic segmentation) of each construction scene.

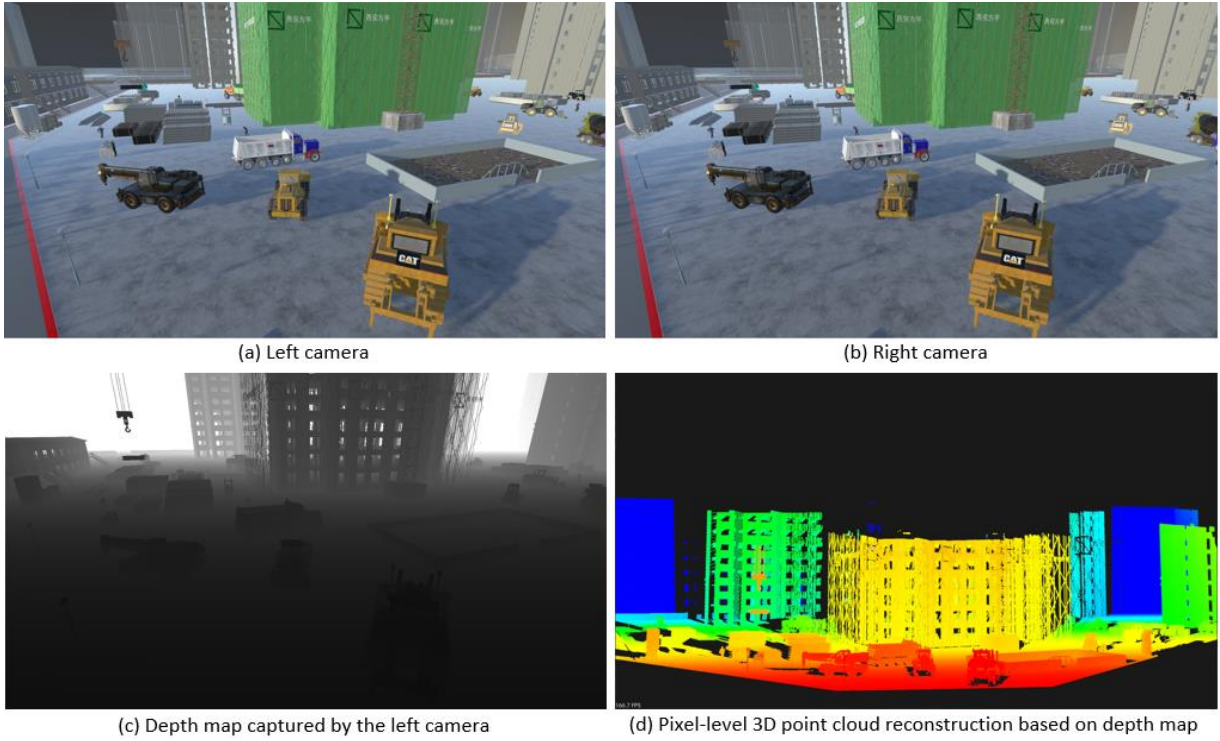

Figure 7. Binocular stereo vision and depth map of the scene.

The performances of data-driven models are often greatly affected by data distribution. Therefore, the dataset generated from scene_A was used for statistical analysis to explore more distribution information about the data. The monitoring area of scene_A was 134×131, and the image number of the Trainval and test sets were 20,000 and 5,000. Figure 7 shows the distribution of per-image instance numbers. Both the distribution of the two subsets roughly conform to the normal distribution. Each image in the Trainval and test set contained an average of 13-14 and 8-9 object instances, respectively. According to one sigma percentage, about 68% of images in the Trainval and test set contained 8-19 and 5-13 instances.

Figure 9 shows the number of different instance sizes calculated by two definitions. The definition of the instance size refers to the COCO dataset [36]: categorizing instances whose bounding box areas (width×height) are less than 32×32, between 32×32 and 96×96, and greater than 96×96 into small, medium, and large object categories, respectively. In this study, we further identified the extra-small/tiny object, i.e., the extra-small instance was identified by bounding box area less than 16×16 while the small instance was between 16×16 and 32×32.

In fact, the actual size of the instance is its visible pixels in the 2D images, not its bounding box area. Therefore, the instance size statistics based on visible pixels were also counted. In general, the statistics results show that instances of medium size or above were dominant in the dataset. In addition, lots of labeled small objects make this dataset ideal for small/tiny object detection research.

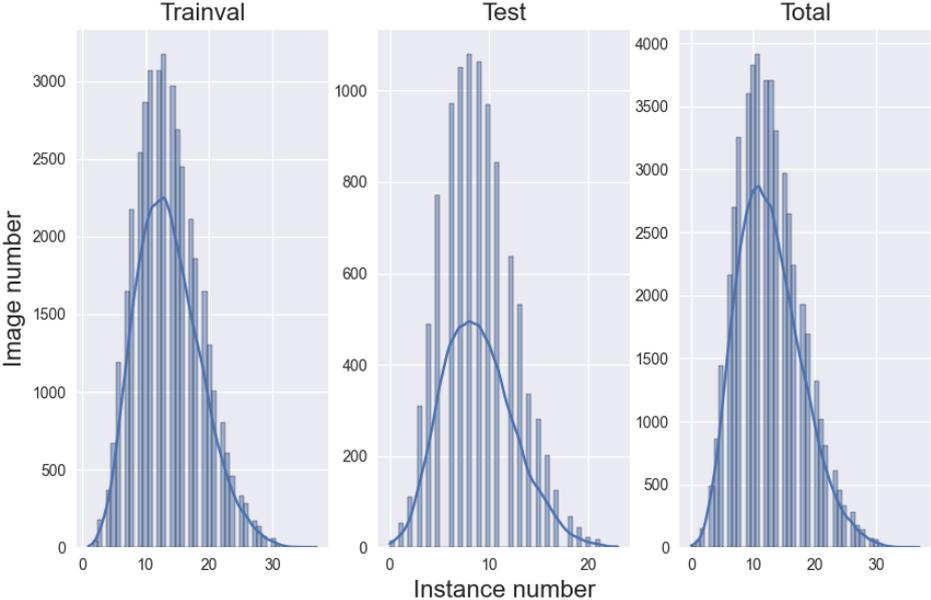

Figure 8. Distribution of per-image instance number.

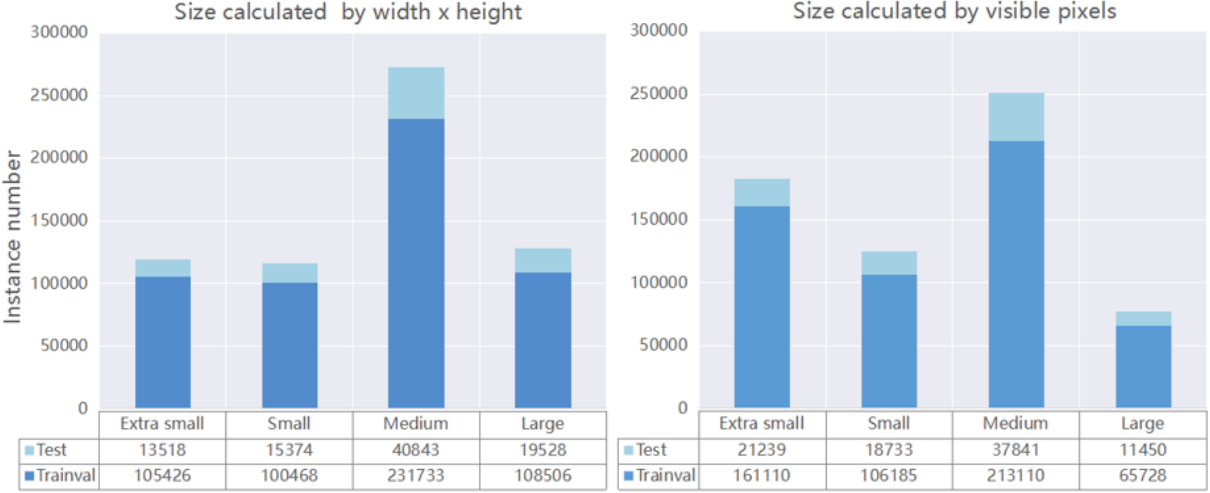

Figure 9. Statistics of instance size.

Figure 10 further explores the instance size distribution within object categories and per-category instance numbers. It can be seen that most small and extra-small instances are workers (more than 75%). For the remaining nine categories, instances above medium size are dominant.

In terms of per-category instance number distribution, the worker and forklift account for more than 47% of the total. The grader has the fewest instances, accounting for only 2.3% of the total. The remaining seven categories account for about 50% of total instances.

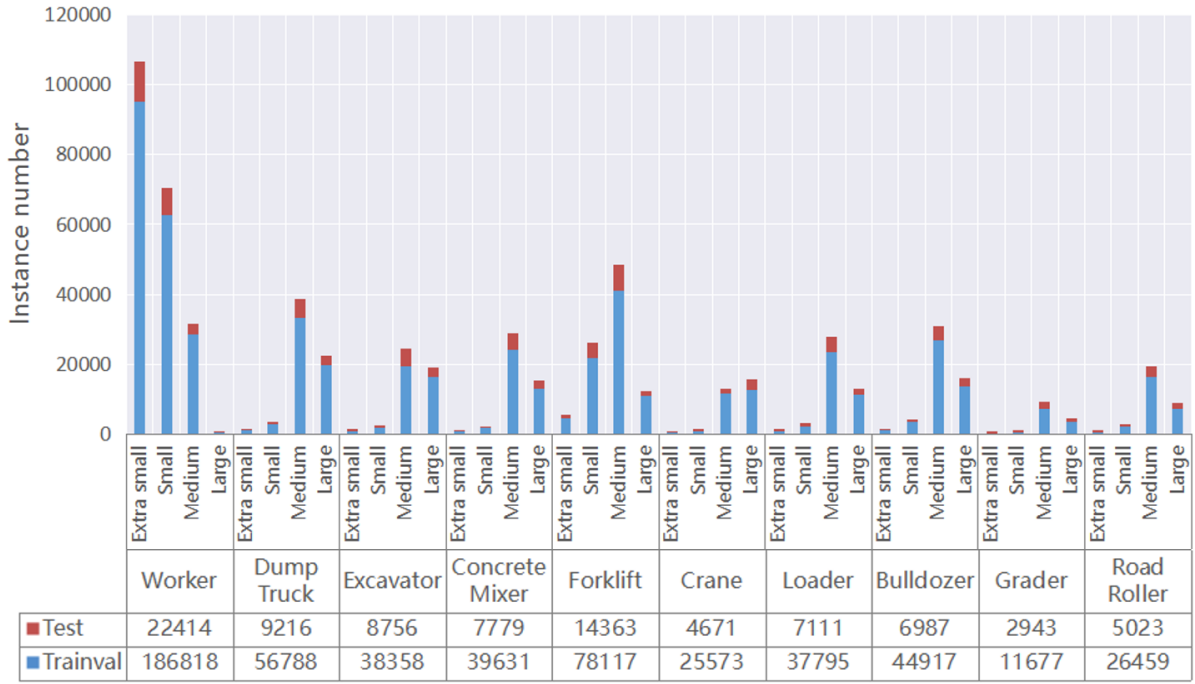

Figure 10. Instance size (width×height) statistics within each object category.

## 5. Benchmark

This section mainly focuses on giving the benchmark performance of some common object detection models on the VCVW-3D dataset. Therefore, there is no depth discussion and analysis of the results and optimal model parameters.

### 5.1 Evaluation metrics

The evaluation metric for 2D object detection was the mean average precision (mAP), which refers to the six metrics defined in COCO [36]: mAP ($mAP_{0.5:0.95}$), $mAP_{0.5}$, $mAP_{0.75}$, $mAP_{small}$, $mAP_{medium}$, and $mAP_{large}$. $AP_{0.5}$ indicates that the intersection over union (IoU) of the predicted bounding box and the ground-truth bounding box is greater than or equal to 0.5, $AP_{0.5:0.95}$ represents the mean value of AP with the IoU from 0.5 to 0.95 (step of 0.05). The small, medium, and large subscripts denote different object sizes defined by the width×height of the bounding box, as described in section 4.

As for 3D object detection, only the monocular image-based models were considered in this study. The mAP and the nuScenes detection score (NDS) defined in [39] were thus adopted. NDS is the weighted average of the mAP, average translation error (ATE), average scale error (ASE), and average orientation error (AOE), as shown in equation (1). For more definition information about the metrics, please refer to [36,39].

$$NDS = \frac{1}{2}mAP + \frac{1}{6}\big(TP\_score(ATE) + TP\_score(ASE) + TP\_score(AOE)\big)$$
$$TP\_score(x) = max(1 - x, 0.0)$$

(1)

**5.2 2D object detection**

For 2D object detection, the Fast R-CNN was used as the representative of the two-stage model, while the SSD and YOLOs were selected as one-stage models. The hyperparameters of each model were kept as their recommended default settings during model training. The selected models were trained, validated, and tested on corresponding data in scene_A of the VCVW-3D dataset. Table 6 shows the benchmark performance of the selected models on scene_A's test data. The YOLOv5-x achieves the best performance in all metrics, especially for small and medium objects, mainly due to its newly proposed mosaic augmentation technology [15]. Table 7 further shows the performance of the YOLOv5-x model for each object category. The YOLOv5-x achieves the lowest AP in the worker category but higher APs in other categories. One of the main reasons is the unbalanced object distribution of different sizes: most workers belong to small objects, while medium and large objects dominate the other categories (Figure 10).

Table 6. Performance of each model on scene_A's test data

| Models | Image size | mAP | $mAP_{0.5}$ | $mAP_{0.75}$ | $mAP_{small}$ | $mAP_{medium}$ | $mAP_{large}$ |
|---|---|---|---|---|---|---|---|
| Faster-RCNN | 640 | 0.479 | 0.610 | 0.538 | 0.006 | 0.404 | 0.717 |

| | | | | | | |
|---|---|---|---|---|---|---|
| SSD | 512 | 0.535 | 0.738 | 0.594 | 0.067 | 0.483 | 0.717 |
| YOLOv3 | 608 | 0.500 | 0.714 | 0.553 | 0.069 | 0.446 | 0.660 |
| YOLOv5-x | 640 | 0.681 | 0.817 | 0.724 | 0.201 | 0.653 | 0.857 |
| YOLOX-s | 640 | 0.485 | 0.645 | 0.530 | 0.079 | 0.414 | 0.673 |

Table 7. Per category performance of the YOLOv5-x model on scene_A's test data.

| Categories | AP | $AP_{0.5}$ | $AP_{0.75}$ | $AP_{small}$ | $AP_{medium}$ | $AP_{large}$ |
|---|---|---|---|---|---|---|
| Worker | 0.346 | 0.666 | 0.313 | 0.301 | 0.772 | 0.638 |
| Dump truck | 0.798 | 0.912 | 0.865 | 0.272 | 0.763 | 0.901 |
| Excavator | 0.708 | 0.832 | 0.761 | 0.179 | 0.688 | 0.884 |
| Concrete mixer | 0.831 | 0.935 | 0.902 | 0.289 | 0.789 | 0.928 |
| Forklift | 0.619 | 0.852 | 0.688 | 0.380 | 0.678 | 0.884 |
| Crane | 0.590 | 0.735 | 0.622 | 0.096 | 0.406 | 0.695 |
| Loader | 0.678 | 0.758 | 0.719 | 0.127 | 0.593 | 0.917 |
| Bulldozer | 0.774 | 0.869 | 0.825 | 0.129 | 0.608 | 0.901 |
| Grader | 0.769 | 0.838 | 0.804 | 0.140 | 0.637 | 0.932 |
| Road roller | 0.696 | 0.777 | 0.740 | 0.099 | 0.593 | 0.889 |

More experiments were carried out to explore the role of virtual synthesis data in real-world applications. The MOCS dataset [43] was selected as the real-world dataset since it has eight identical categories to the VCVW-3D. The available labeled data of the MOCS were only the training set (19,404 images) and the validation set (4,000 images). The experiments were designed as follows: 1) train three models using scene_A's training data, MOCS's validation data, and scene_A's training data + MOCS's validation data, respectively; 2) test these three models on MOCS's training data.

As shown in Table 8, the model trained only on scene_A's training data achieves a terrible performance on real-world data, indicating that real-world images are irreplaceable and pure virtual data cannot completely replace real-world data. Besides, the model trained on scene_A's training data + MOCS's validation data achieves the best performance, and AP metrics of each category are close to or even exceed the benchmark performance of the MOCS [43]. Note that the original benchmark performance of MOCS was trained on a large real-world dataset (19,404 images) and tested on a real-world dataset of 18,264 images. The test results in this study and MOCS's original experiment are comparable because MOCS's training and test datasets are in the same distribution and similar size. Therefore, the superior result achieved in this study reveals that virtual synthetic images can help achieve similar or even better performance than those trained on large real-world datasets by combining only a small amount of real-world data. Though this finding is not first proposed in this study [7], the result proves the finding is applicable in construction.

Table 8. Per category performance of the YOLOv5-x on the MOCS's training data.

| Training data | Categories | AP | $AP_{0.5}$ | $AP_{0.75}$ | $AP_{small}$ | $AP_{medium}$ | $AP_{large}$ |
|---|---|---|---|---|---|---|---|
| Only scene_A's training data | All | 0.004 | 0.015 | 0.002 | 0.000 | 0.003 | 0.005 |
| Only MOCS's validation data | All | 0.537 | 0.721 | 0.577 | 0.148 | 0.354 | 0.703 |
| Scene_A's training data + MOCS's validation data | Worker | 0.499 | 0.804 | 0.517 | 0.279 | 0.563 | 0.742 |
| | Dump truck | 0.519 | 0.709 | 0.548 | 0.190 | 0.411 | 0.726 |
| | Excavator | 0.718 | 0.892 | 0.778 | 0.173 | 0.534 | 0.830 |
| | Concrete mixer | 0.481 | 0.708 | 0.498 | 0.161 | 0.392 | 0.692 |
| | Crane | 0.466 | 0.658 | 0.502 | 0.077 | 0.282 | 0.570 |
| | Loader | 0.531 | 0.702 | 0.582 | 0.065 | 0.307 | 0.680 |
| | Bulldozer | 0.662 | 0.830 | 0.718 | 0.093 | 0.392 | 0.779 |
| | Road roller | 0.735 | 0.870 | 0.790 | 0.154 | 0.464 | 0.853 |
| | **Mean** | 0.576 | 0.772 | 0.617 | 0.149 | 0.418 | 0.734 |

**5.3 Monocular 3D object detection**

Since there was no relevant real-world labeled dataset of 3D object detection in the construction field, the 3D object detection models in this study were totally based on the data in scene_A of the VCVW-3D. As shown in Table 9, two advanced 3D object detection models, PGD and FCOS3D, were trained and tested in nuScenes data format. According to the results, the FCOS3D model achieves the best performance in both mAP and NDS metrics. Figure 11 shows the 2D projection of 3D bounding boxes predicted by the FCOS3D model. Besides, unlike 2D object detection, 3D object detection models in this section seem biased towards categories containing more small objects (e.g., the worker and forklift). Especially for the PGD model, the worker and forklift categories tend to achieve higher AP values. One of the possible reasons is that large objects are more likely to be occluded by their surroundings. Though some space collision avoidance measures were adopted in data synthesis, occlusions between objects still occurred when projecting the current scene onto a 2D image (3.3). Therefore, it is still tricky for current advanced 3D object detection models to predict the specific position and size of the occluded part in space. More experiments and studies are thus worth being explored in this direction.

Table 9. Per category performance of 3D object detection models on scene_A's test data.

| Models | Categories | AP | ATE | ASE | AOE | mAP | NDS |
| --- | --- | --- | --- | --- | --- | --- | --- |
| PGD | Worker | 0.646 | 0.323 | 0.752 | 1.619 | 0.457 | 0.351 |
| | Dump truck | 0.529 | 0.437 | 0.763 | 1.524 | | |
| | Excavator | 0.358 | 0.623 | 0.723 | 1.535 | | |
| | Concrete mixer | 0.629 | 0.438 | 0.786 | 1.571 | | |
| | Forklift | 0.618 | 0.409 | 0.746 | 1.642 | | |
| | Crane | 0.181 | 0.820 | 0.797 | 1.804 | | |
| | Loader | 0.413 | 0.449 | 0.782 | 1.53 | | |
| | Bulldozer | 0.401 | 0.611 | 0.686 | 1.609 | | |
| | Grader | 0.413 | 0.484 | 0.793 | 1.636 | | |
| | Road roller | 0.385 | 0.506 | 0.717 | 1.491 | | |
| FCOS3D | Worker | 0.577 | 0.352 | 0.328 | 1.731 | 0.465 | 0.439 |
| | Dump truck | 0.513 | 0.435 | 0.213 | 2.032 | | |
| | Excavator | 0.414 | 0.552 | 0.355 | 1.912 | | |
| | Concrete mixer | 0.608 | 0.412 | 0.219 | 1.988 | | |
| | Forklift | 0.595 | 0.400 | 0.285 | 1.981 | | |
| | Crane | 0.193 | 0.834 | 0.476 | 1.886 | | |
| | Loader | 0.453 | 0.372 | 0.113 | 2.049 | | |
| | Bulldozer | 0.392 | 0.586 | 0.232 | 1.975 | | |
| | Grader | 0.509 | 0.447 | 0.224 | 2.043 | | |
| | Road roller | 0.398 | 0.513 | 0.258 | 1.794 | | |

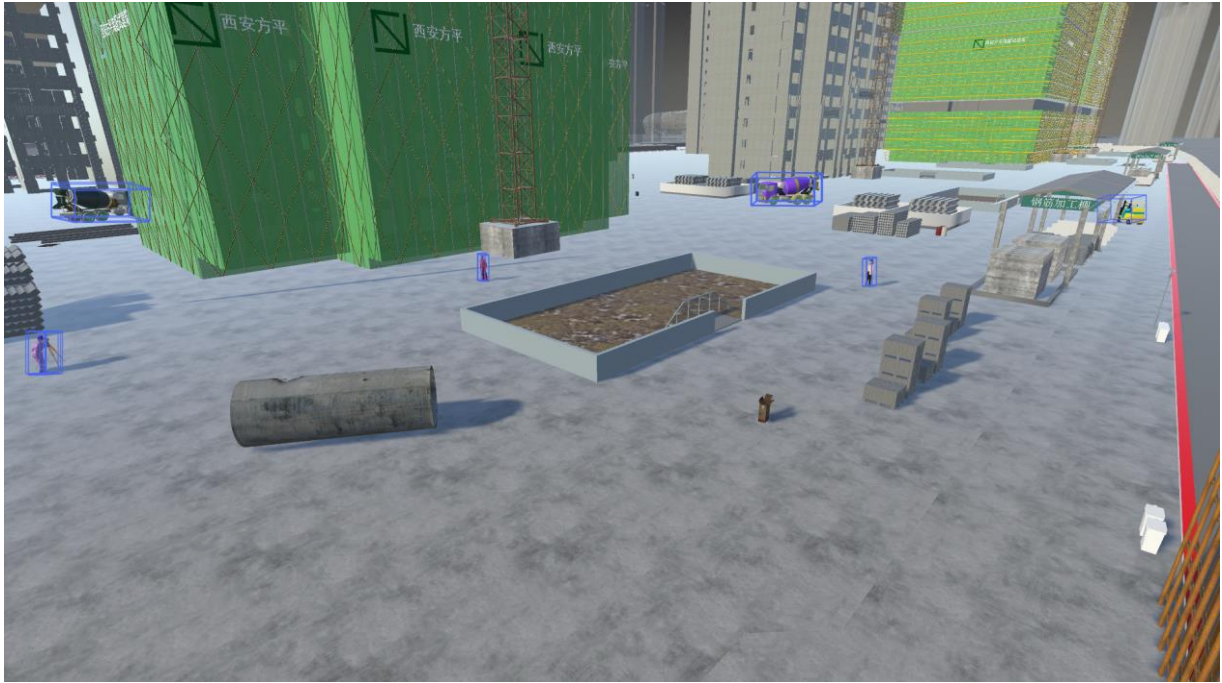

Figure 11. 3D bounding box predictions using the FCOS3D model.

## 6. Discussion

The VCVW-3D dataset is expected to have the following impact and contributions:

- First, the VCVW-3D dataset might bring substantial economic benefits to 3D CV development in construction by reducing data access costs and speeding up the development and verification of prototype applications. It is well known that data acquisition and annotation are tedious and time-consuming [6]. With the help of virtual technology, a great number of data with annotations can be automatically generated only using a small number of relevant prefabricated assets, which frees researchers from the complicated and time-consuming work of data collection, processing, annotation, and verification. It is also proven that virtual data can help achieve a better model by combining only a small amount of real-world data, which is suitable for the rapid development and verification of prototype models.

- Secondly, the VCVW-3D dataset may stimulate more 3D CV research based on 2D images in construction. Compared with pure 2D datasets, the VCVW-3D dataset annotated multiple spatial information about objects, such as their center point, size, and orientation angle in 3D space, making it possible for the trained model to perceive objects' spatial information from 2D images. More advanced space-awareness applications in construction based on 2D cameras thus become explorable, such as high-altitude operations judgment, safety distance estimation, relative space position recognition, etc. Currently, monocular 2D surveillance cameras are universally installed on many construction sites. Therefore, 3D information prediction based on monocular 2D camera data is a more practical and economical choice than deploying expensive LiDAR or other depth sensors.

- Finally, the VCVW-3D dataset might promote the extensive development of 3D CV in the construction area. Data is the research foundation of data-driven models. Due to the lack of 3D annotated data, the development of 3D CV in construction is far behind the general field. In addition to monocular/binocular 3D object detection, monocular/binocular depth estimation, scene reconstruction, and other multimodal (monocular/binocular 2D image + depth map) tasks can also be studied, validated, and applied in construction based on the VCVW-3D dataset. The VCVW-3D dataset is also expected to set an example to appeal to more excellent public 3D datasets in construction.

However, the VCVW-3D dataset still has some limitations. First, the VCVW-3D contains only three types of 3D information: 3D bounding box, depth map, and binocular stereo vision. Therefore, more relevant 3D annotation information, such as the virtual LiDAR point cloud of the scene, 3D key point annotation of workers (3D posture) and rotatable equipment (excavators, cranes, etc.). Secondly, the VCVW-3D only includes objects of workers and construction vehicles and lacks objects of common construction materials. Thus, expanding the coverage of object categories is another work to be solved in the future.

## 7. Conclusion

This study generates and publicly releases a virtual image dataset with several 2D and 3D annotations named VCVW-3D, which covers 15 construction scenes and involves ten categories of workers and large construction vehicles. Each scene contributes 20,000 and 5,000 images of 1920×1080 for the Trainval and test sets, annotated with 2D/3D bounding boxes, 2D semantic/instance segmentation, and depth maps. The VCVW-3D also has other characteristics of multi-randomness, multi-viewpoint, and stereo vision. As one of the critical information of 3D position annotation, object orientation is tentatively discussed and defined in this study for the first time. A series of 2D and monocular 3D object detection models are subsequently trained and evaluated under the scene_A of the VCVW-3D to provide a benchmark reference for subsequent research. The benchmark results also show that pure virtual data cannot completely replace real-world data but indeed help to obtain similar or even better models when only combining a small number of real-world data. The VCVW-3D is expected to bring considerable economic benefits and practical significance by reducing the costs of data acquisition, prototype development, and exploration for advanced space-awareness applications. Besides object detection, the VCVW-3D dataset can also be used for many other 2D/3D CV studies, which might promote the comprehensive development of CV in construction, especially in 3D applications.

## 8. Data availability statement

The VCVW-3D dataset can be publicly accessed at https://github.com/dyxm/VCVW-3D.

**Acknowledgment**

Acknowledgment